%% file: main.tex
\documentclass{article}
\usepackage{framelite_preprint,times}

\usepackage[T1]{fontenc}
\usepackage[utf8]{inputenc}
\usepackage{amsmath,amssymb,bm}
\usepackage{graphicx}
\usepackage{booktabs,multirow,tabularx,makecell}
\usepackage{xcolor}
\usepackage{microtype}
\usepackage{placeins}
\usepackage[pagebackref=true,breaklinks=true,letterpaper=true,colorlinks,bookmarks=false]{hyperref}

\graphicspath{{figs/}}
\newcommand{\method}{FrameLite}

\hypersetup{
  pdftitle={FrameLite: Decision-Oriented Aesthetic Cropping with a Compact Vision--Language Model}
}
\hypersetup{pdfauthor={Jianze Li, Kai Liu, Ziqing Zhang, Weihang Zhang, Yong Guo, Linghe Kong, Yulun Zhang}}

\title{\raggedright
  FrameLite: Decision-Oriented\\
  Aesthetic Cropping with a Compact\\
  Vision--Language Model
}

\author{
  \textbf{Jianze Li$^{1}$,\enspace
  Kai Liu$^{1}$,\enspace
  Ziqing Zhang$^{1}$,\enspace
  Weihang Zhang$^{2}$,}\\
  \textbf{Yong Guo$^{1}$,\enspace
  Linghe Kong$^{1}$,\enspace
  Yulun Zhang$^{1}$}\thanks{Corresponding author: Yulun Zhang, yulun100@gmail.com.}\\[3pt]
  \textsuperscript{1}Shanghai Jiao Tong University\\
  \textsuperscript{2}Institute of Media Technology and Experience Design, Huawei
}
\date{}

\begin{document}
\flushbottom
\setcounter{footnote}{1}
\maketitle
\lhead{\small\normalfont FrameLite: Decision-Oriented Aesthetic Cropping with a Compact Vision--Language Model}


\vspace{-6mm}
\begin{abstract}
Aesthetic image cropping improves composition by balancing subjects,
context, and image boundaries. Large vision--language models bring
semantic knowledge to this task, but their scale limits lightweight
photography assistance. We propose \method{}, a compact cropping model
built on Qwen3.5-0.8B. Our approach is motivated by the observation that
a small model can generate better crops than its first answer suggests.
We develop three-stage training to turn this potential into stronger
direct predictions. Spatial and composition training establishes
localization and framing preferences. Quality-Guided Online Supervision
(QOS) then learns the highest-quality complete answers among the current
policy's generated candidates. Greedy-Referenced Optimization (GRO)
further refines the policy using positive and negative quality differences
relative to its current greedy answer. Both online stages use
annotation-derived feedback, while inference returns one coordinate
answer without search or an external scorer. Experiments show that \method{} outperforms the compared large-VLM
croppers across public benchmarks with a substantially smaller backbone. Controlled studies
confirm that selecting high-quality training answers improves the final
model beyond online sampling alone.
We will release our code and model weights at
\url{https://github.com/JianzeLi-114/FrameLite}.
\end{abstract}
\vspace{-3mm}

\section{Introduction}
\label{sec:introduction}

Aesthetic image cropping selects a visually coherent region by adjusting
image boundaries, removing distractions, and balancing subjects with
their surroundings. Beyond locating salient objects, it determines how
much context to retain, where to leave space, and which visual
relationships to preserve. Multiple windows can therefore form
reasonable compositions~\citep{jia2022rethinking}. Progress spans view evaluation and proposal
learning~\citep{chen2017vfn,wei2018view}, grid
anchors~\citep{zeng2019grid,zeng2020grid}, and composition-aware
or spatial--semantic
modeling~\citep{hong2021composing,zhang2022hcic,su2024s2cnet}.
Feature extrapolation, professional-composition transfer, and global
assessment provide further ways to learn
framing~\citep{liu2023unic,hong2024gencrop,zhang2026procrop,yang2026gafic}.
These advances support automatic reframing and motivate stronger
semantic reasoning about subject placement and photographic composition.

Vision--language models (VLMs) connect cropping with instruction
following, aesthetic knowledge, and contextual
examples~\citep{zhong2022clipcrop,sheng2025instructcrop,du2026venus,lee2025cropper}.
Current methods also incorporate compositional reasoning, multimodal
guidance, and capture-time
assistance~\citep{dong2026crop,you2025photoframer,li2026shuttermuse}.
Their deployment requirements can nevertheless be demanding for everyday
photography. Cropper uses an external Gemini service with retrieval and
iterative refinement~\citep{lee2025cropper}; InstructCrop's released
implementation adopts
LLaVA-1.6-7B~\citep{sheng2025instructcrop,instructcropcode2025};
Venus-Q and CROP also use 7B-class
backbones~\citep{du2026venus,dong2026crop}.
These systems offer valuable explanations and interactions. A person
composing a photograph on a phone may need one immediate, actionable
suggestion~\citep{li2025spas}. Large model weights place pressure on device memory, while
cloud access and repeated generation add dependencies and response
costs. This motivates a compact VLM that can make a strong framing
decision through a short, self-contained prediction path. The practical
challenge is to express semantic judgments as precise boundaries within
this compact model, while keeping the interaction as simple as a single
actionable framing suggestion for the photographer.

\begin{figure}[t]
  \centering
  \includegraphics[width=\linewidth]{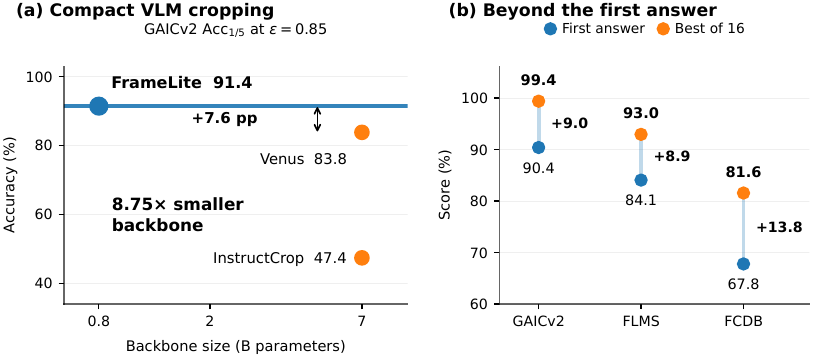}
  \vspace{-4mm}
  \caption{\textbf{Better first answers with a compact model.}
  (a) GAICv2 $\mathrm{Acc}_{1/5}$ at $\epsilon=0.85$ versus nominal VLM
  backbone size; the horizontal line marks \method{}.
  (b) First-answer and best-of-16 results from the same early supervised
  Qwen model. Best of 16 selects the highest-scoring sampled crop using
  human reference annotations. Scores are percentages of the GAICv2
  $\mathrm{Acc}_{1/5}^{0.85}$ acceptance rate, FLMS maximum-reference IoU,
  and FCDB single-reference IoU.}
  \label{fig:intro_overview}
  \vspace{-3mm}
\end{figure}

Our preliminary experiments reveal a useful opportunity: a small model
trained with grounding and composition supervision can generate better
crops than its first answer suggests. Choosing the best of sixteen
samples using human references substantially improves quality across
the evaluated benchmarks (Figure~\ref{fig:intro_overview}b).
This gap motivates target selection: the model must learn which of its
plausible compositions best reflects the annotated preference. Test-time search would pass exploration costs to
the user, while fixed targets leave the model's current alternatives
unexamined. This raises a central question: \emph{how can a compact model learn
from its own alternatives to make a stronger first-answer cropping decision for the same input photograph?}

We propose \method{}, a direct aesthetic cropping model built on
\textbf{Qwen3.5-0.8B}~\citep{qwen2026}. Spatial and composition training
connects localization, candidate-quality learning, and complete-frame
supervision. \textbf{Quality-Guided Online Supervision (QOS)} then
learns the highest-reward answer among the policy's greedy and sampled
crops, adapting reward-selected learning~\citep{dong2023raft} to its
evolving outputs. We further propose \textbf{Greedy-Referenced
Optimization (GRO)}, our crop-specific use of self-critical
optimization~\citep{rennie2017scst}, to learn the quality differences
between sampled crops and the current greedy answer.
Training thus moves from learning what to frame, to finding better
answers, to refining their likelihoods. Deployment retains the same compact
model and a single coordinate answer, with no candidate selection.

Across public cropping benchmarks, \method{} delivers leading
performance among the compared VLM croppers with a fraction of their
nominal backbone size (Figure~\ref{fig:intro_overview}a).
With the same number of generated training candidates, learning the
best answers yields a stronger final model than learning random ones,
even after the same subsequent GRO training. These results connect
compact models with accurate direct framing for lightweight
photography assistance.

Our contributions connect compact crop prediction with progressive decision learning:
\begin{itemize}
  \item \textbf{A compact aesthetic cropping model.}
  We propose \method{}, a sub-billion-parameter vision--language model
  that combines spatial understanding and composition learning to
  predict a complete crop directly from the input image.
  \item \textbf{Quality-Guided Online Supervision (QOS).}
  We learn refreshed, high-quality complete answers from the current
  policy, connecting annotated preferences to its generated alternatives.
  \item \textbf{Progressive refinement with GRO.}
  We couple winner imitation with signed, greedy-referenced quality
  feedback to refine the policy within the same compact architecture.
\end{itemize}

\section{Related Work}
\label{sec:related_work}

\subsection{Aesthetic Cropping and Photographic Assistance}

Learning-based cropping models compositional preferences through candidate
scoring, ranking, and direct prediction. VFN, VEN, and VPN learn
view evaluation and proposals with pairwise supervision~\citep{chen2017vfn,wei2018view},
while grid anchors reduce the candidate space~\citep{zeng2019grid,zeng2020grid}.
Attention and subject-aware models guide crop generation and
refinement~\citep{wang2017deepcropping,yang2023subject}. ASM-Net, CACNet,
HCIC, and S$^2$CNet model salient-region placement, composition rules,
human-relative structure, and object
relationships~\citep{tu2020asm,hong2021composing,zhang2022hcic,su2024s2cnet}.
CGS and TransView capture relations among candidates and across crop
boundaries~\citep{li2020cgs,pan2021transview}; set prediction accommodates
diverse compositions~\citep{jia2022rethinking}, and spatial-aware features
with rank consistency extend comparative supervision to unlabeled
images~\citep{wang2023spatialrank}. Feature extrapolation, outpainting
supervision, professional-composition retrieval, and global feature
alignment provide complementary sources of visual context, candidate comparisons,
and photographic composition knowledge~\citep{liu2023unic,hong2024gencrop,zhang2026procrop,yang2026gafic}.

Vision--language methods connect photographic quality with semantic
intent. VILA learns aesthetic representations from image--comment
pairs~\citep{ke2023vila}, while AesExpert and PhotoEye support aesthetic
instruction following and photographic
critique~\citep{huang2024aesexpert,qi2025photographer}. ClipCrop uses CLIP
representations for text- or image-conditioned
cropping~\citep{radford2021clip,zhong2022clipcrop}; InstructCrop and Venus
learn cropping instructions and rationale-supported
prediction~\citep{sheng2025instructcrop,du2026venus}. Cropper combines
retrieved examples with iterative refinement~\citep{lee2025cropper},
whereas CROP aligns compositional reasoning with expert
preferences~\citep{dong2026crop}. Beyond crop prediction, smart
point-and-shoot photography offers camera-pose guidance~\citep{li2025spas},
and PhotoFramer and ShutterMuse provide multimodal composition
instructions and capture-time
assistance~\citep{you2025photoframer,li2026shuttermuse}.

\subsection{Reward-Guided Learning and Policy Optimization}

Reward-guided learning uses generated outputs to construct
training supervision.
RAFT samples candidate responses, selects high-reward examples,
and fine-tunes the model on the selected
outputs~\citep{dong2023raft}.
Cropping-specific reinforcement learning includes A2-RL and Fast A3RL,
which learn sequential window adjustments under aesthetic
feedback~\citep{li2018a2rl,li2019fasta3rl}.
For sequence generation, self-critical sequence training (SCST)
uses the model's own inference output as a reward baseline
and updates sampled sequences according to their relative
rewards~\citep{rennie2017scst}.
These approaches provide distinct ways to exploit outcome
quality, through selected-output imitation or reward-weighted
updates to the learned policy.

\setlength{\parskip}{5pt plus 1.5pt minus 0.7pt}
\section{Method}
\label{sec:method}

\begin{figure}[t]
  \centering
  \includegraphics[width=\linewidth]{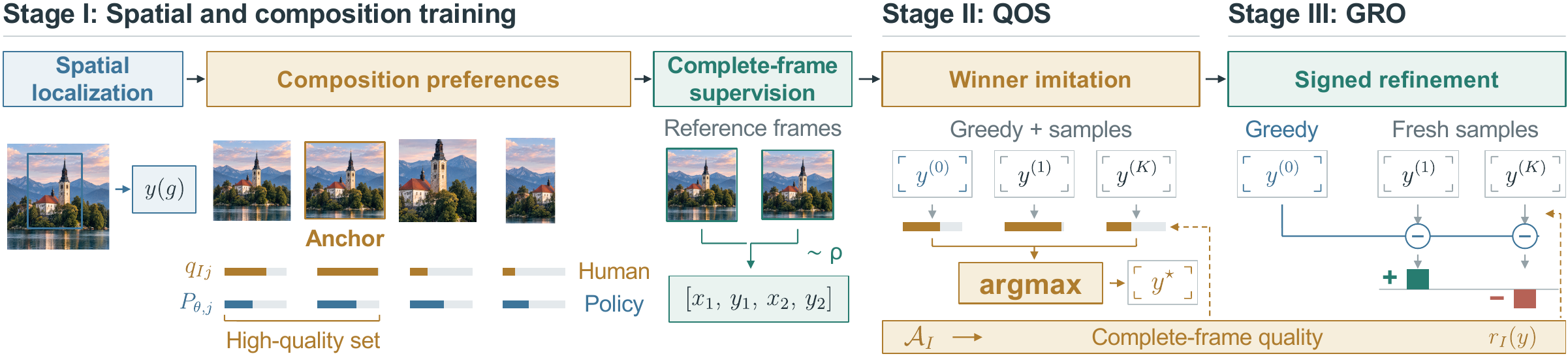}
  \vspace{-5mm}
  \caption{\textbf{Three-stage training of \method{}.}
  For annotated crop $j$ of image $I$, $q_{Ij}$ and $P_{\theta,j}$ are human
  and policy preference probabilities; $\rho$ gives reference-sampling weights.
  $y(g)$ serializes region $g$ as normalized corner coordinates $(x_1,y_1,x_2,y_2)$.
  QOS imitates the best answer $y^\star$ among greedy $y^{(0)}$ and $K$ samples
  $y^{(1:K)}$; GRO refines fresh samples relative to greedy. Both use quality
  $r_I(y)$ defined by human crop annotations $\mathcal A_I$ associated with each training image.}
  \label{fig:method}
  \vspace{-2mm}
\end{figure}

\begin{figure}[t]
  \centering
  \includegraphics[width=\linewidth]{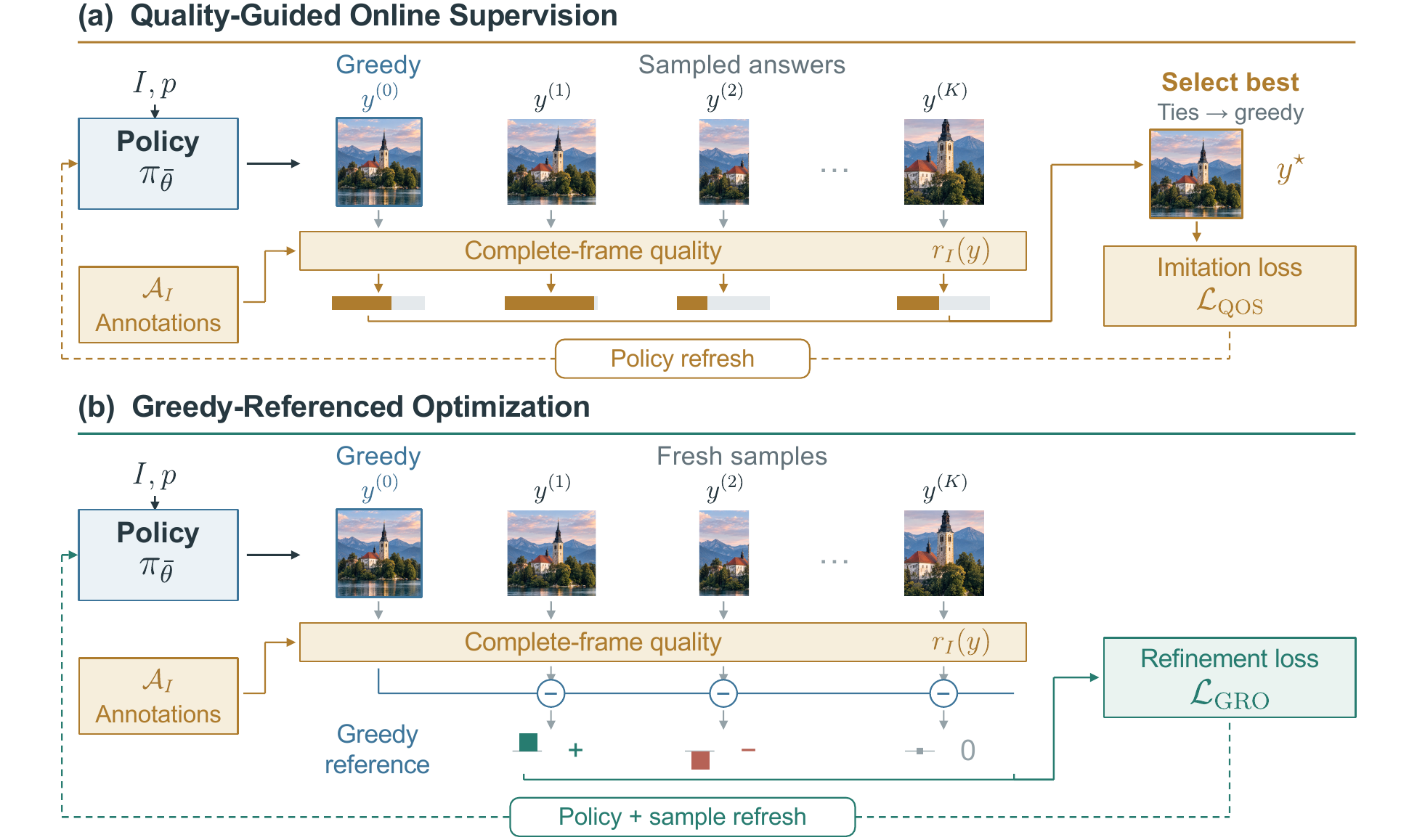}
  \vspace{-5mm}
  \caption{\textbf{Online supervision and refinement.}
  Given image $I$ and prompt $p$, policy $\pi_{\bar\theta}$ with current parameters
  $\bar\theta$ generates greedy answer $y^{(0)}$ and independent samples
  $y^{(j)}$, $j=1,\ldots,K$. Annotations $\mathcal A_I$ define quality $r_I(y)$.
  (a) $\mathcal L_{\mathrm{QOS}}$ increases the likelihood of the best answer
  $y^\star$, favoring greedy on ties. (b) $\mathcal L_{\mathrm{GRO}}$ weights
  sampled answers' log-likelihoods by signed quality differences from greedy.
  Dashed loops refresh the policy and candidates after each update.}
  \label{fig:online_training}
  \vspace{-2mm}
\end{figure}

\method{} uses three-stage decision learning to locate important content,
assess alternative framings, and commit to one answer. \textbf{Spatial and composition training}
first learns from annotated regions, crop preferences, and complete
reference frames (Figure~\ref{fig:method}).
\textbf{Quality-Guided Online Supervision (QOS)} then selects a
high-quality answer from the current policy's greedy and sampled crops
(Figure~\ref{fig:online_training}(a)). \textbf{Greedy-Referenced Optimization (GRO)}
uses each new sample's quality relative to the current greedy answer
(Figure~\ref{fig:online_training}(b)). All three stages improve the same compact policy for direct crop prediction
from a single greedy coordinate answer at inference.

\subsection{Formulation and Model}
\label{sec:overview}

A crop's quality depends on the joint placement of its boundaries.
We represent it as a normalized box in
$\mathcal B=\{(x_1,y_1,x_2,y_2):0\leq x_1<x_2\leq1,\;
0\leq y_1<y_2\leq1\}$. Given image $I$ and fixed prompt $p$, the policy
generates a coordinate sequence $y$:
\begin{equation}
\pi_\theta(y\mid I,p)
=\prod_{t=1}^{T(y)}\pi_\theta(y_t\mid I,p,y_{<t}).
\label{eq:policy}
\end{equation}
A parser $\mathcal P$ maps $y$ to a complete box or invalid symbol
$\bot$. We use Qwen3.5-0.8B~\citep{qwen2026} with a frozen visual encoder.
Stage I adapts its visual merger and language LoRA~\citep{hu2022lora}; subsequent stages
update language adapters. The training distribution $\mathcal D$ contains
images and crop annotations $\mathcal A_I$, with the chosen example
weights. Annotations provide supervision and quality feedback, without
being supplied to the prediction model.

\subsection{Stage I: Spatial and Composition Training}
\label{sec:spatial_composition}

\paragraph{Spatial localization.}
A subject's position and extent organize the surrounding composition.
We therefore first teach the policy to locate important content through
referring-expression localization~\citep{yu2016context}. For an expression $e$ and annotated
region $g$, let $y(g)$ serialize its coordinates. The loss averages
negative log likelihood over valid target tokens in a training batch:
\begin{equation}
\mathcal L_{\mathrm{loc}}
=-\frac{1}{|\mathcal T_{\mathrm{loc}}|}
 \sum_{(i,t)\in\mathcal T_{\mathrm{loc}}}
 \log\pi_\theta(y(g_i)_t\mid I_i,e_i,y(g_i)_{<t}),
\label{eq:grounding}
\end{equation}
where $\mathcal T_{\mathrm{loc}}$ indexes those tokens. The region target in Figure~\ref{fig:method}
teaches spatial grounding; it does not prescribe a tight subject box as
the final aesthetic crop. The remaining training steps learn how to place the final frame
around the located subject and its surrounding context.

\paragraph{Composition preferences.}
Several windows may preserve the same subject while differing in context,
balance, and empty space; learning one reference leaves their relative
quality unused. Given rated annotations
$\mathcal A_I=\{(g_j,u_j)\}_{j=1}^{M_I}$, we therefore align the policy's
preferences over complete answers with human scores. Let
$z_{\theta,j}=\log\pi_\theta(y(g_j)\mid I,p)$ and define the two
preference distributions over the annotated candidate frames of the same input image:
\begin{equation}
q_{Ij}=\frac{\exp(\widetilde u_j/\tau)}
                  {\sum_k\exp(\widetilde u_k/\tau)},
\qquad
P_{\theta,j}=\frac{\exp(z_{\theta,j})}
                     {\sum_k\exp(z_{\theta,k})},
\label{eq:preference}
\end{equation}
where $\widetilde u_j$ is a normalized annotation score and $\tau$
controls the concentration of the target distribution.
The model probabilities are normalized over the full annotated
candidate list so that the relative likelihoods of alternative framings
can be compared directly.

A good target should provide a concrete answer without discarding the
fact that several answers may be acceptable. We select an anchor $g_a$
as a quality-weighted IoU medoid within the high-quality subset
$\mathcal H_I$, and combine three complementary signals:
\begin{equation}
\mathcal L_{\mathrm{comp}}
=\mathbb E_I\!\left[
 \underbrace{-z_{\theta,a}}_{\text{anchor}}
 \;\underbrace{-\sum_jq_{Ij}\log P_{\theta,j}}_{\text{relative quality}}
 \;\underbrace{-\log\!\left(\sum_{j\in\mathcal H_I}P_{\theta,j}\right)}_{\text{high-quality set}}
\right].
\label{eq:composition}
\end{equation}
The anchor supplies a concrete answer, the listwise term aligns quality
preferences, and the set term favors multiple acceptable compositions
without fixing their probability allocation. Together, these signals
adapt complete-answer learning to the ambiguity of aesthetic framing.

\paragraph{Complete-frame supervision.}
Relative preferences alone specify which annotated crops should rank
higher; direct prediction also requires assigning likelihood to full
coordinate answers. We next consolidate this behavior using the
available reference frames and their fixed annotation-derived weights
$\rho_{Ij}\geq0$, with $\sum_j\rho_{Ij}=1$ for each training image:
\begin{equation}
\mathcal L_{\mathrm{ann}}
=-\mathbb E_{(I,\mathcal A_I)\sim\mathcal D}
 \left[\sum_j\rho_{Ij}\log\pi_\theta(y(g_j)\mid I,p)\right].
\label{eq:annotation}
\end{equation}
A single reference receives unit weight; multiple references can express
quality or acceptance preferences. Sampling complete references under
these weights preserves alternative framings, instead of creating a new
box by averaging their boundaries. This supervision also accommodates
images with a final crop but no candidate ranking. Spatial localization,
preference learning, and full-answer supervision are successive parts
of Stage I, yielding the initial policy $\pi_{\theta_{\mathrm{init}}}$.

\subsection{Stage II: Quality-Guided Online Supervision (QOS)}
\label{sec:qos}

Annotated frames do not reveal which alternatives a policy generates
when making its own decisions. We propose Quality-Guided Online
Supervision (QOS) to bridge this gap by selecting a complete answer from
the current policy under an annotation-defined quality criterion.
Refreshed during training, the target connects the model's current
outputs to human framing preferences.

To compare complete frames, let $R(I,b;\mathcal A_I)$ evaluate a crop
using its training annotations. A single accepted frame supports endpoint
agreement; several acceptable frames support best-reference matching;
and rated alternatives support quality-weighted agreement. We assign a
sequence reward after its complete coordinates have been generated:
\begin{equation}
 r_I(y)=
 \begin{cases}
 R(I,\mathcal P(y);\mathcal A_I), & \mathcal P(y)\in\mathcal B,\\
 r_{\mathrm{invalid}}, & \mathcal P(y)=\bot,
 \end{cases}
\label{eq:quality}
\end{equation}
where $r_{\mathrm{invalid}}$ is below any valid reward. The same fixed
criterion guides both online stages, so target selection and policy
refinement optimize a consistent notion of complete-frame quality.

At the start of an update, let $\bar\theta$ denote the current parameters.
The policy generates its greedy answer and $K$ independent alternatives, then
selects the highest-reward complete answer:
\begin{equation}
\begin{aligned}
 y^{(0)}&=\operatorname{Greedy}(\pi_{\bar\theta},I,p),
 & y^{(j)}&\sim\pi_{\bar\theta}(\cdot\mid I,p),\\
 j^\star&=\arg\max_{j\in\{0,\ldots,K\}}r_I(y^{(j)}),
 & y^\star&=y^{(j^\star)}.
\end{aligned}
\label{eq:winner}
\end{equation}
Including the greedy answer gives the selection a natural fallback:
$r_I(y^\star)\geq r_I(y^{(0)})$ within each generated set. Ties favor the
greedy answer. Thus, exploration supplies alternatives while the quality
criterion decides whether any is a better target than the current
first response. This relation concerns target selection, not a guarantee
that every parameter update improves greedy quality.

QOS then increases the likelihood of the selected answer:
\begin{equation}
\mathcal L_{\mathrm{QOS}}(\theta;\bar\theta)
=-\mathbb E_{(I,\mathcal A_I)\sim\mathcal D}
 \mathbb E_{y^{(1:K)}\sim\pi_{\bar\theta}}
 \left[\log\pi_\theta(y^\star\mid I,p)\right].
\label{eq:qos}
\end{equation}
Candidates and selection are held fixed during differentiation, with
the update evaluated at $\theta=\bar\theta$. The updated policy supplies
the next batch of candidates. This use of reward-selected
learning~\citep{dong2023raft} couples target discovery to the policy's
actual decisions, while human annotations keep selection tied to
composition quality. Each image contributes one selected complete
answer; QOS does not explicitly encode how much the unselected answers
improve on or fall below the current prediction for the same input image at the current stage of policy learning.

\subsection{Stage III: Greedy-Referenced Optimization (GRO)}
\label{sec:gro}

Greedy-Referenced Optimization (GRO) extends winner imitation to feedback
on all sampled frames, evaluating each complete framing decision relative
to the answer the policy would currently return.

A crop's absolute score does not indicate improvement over the current
decision. The greedy answer provides an image-specific, policy-dependent
reference. For fresh samples from the current policy, we define a signed
quality advantage for each sampled coordinate answer:
\begin{equation}
A_j=r_I(y^{(j)})-r_I(y^{(0)}),\qquad j=1,\ldots,K.
\label{eq:advantage}
\end{equation}
A positive advantage identifies a sampled frame that is better under
the training criterion; a negative advantage identifies an inferior
alternative. Because the reference is recomputed from the current
policy, the comparison follows its evolving framing ability rather
than a fixed historical answer.

We use the self-critical sequence-training
estimator~\citep{rennie2017scst} to optimize the quality of complete
coordinate answers under this image-specific reference:
\begin{equation}
\mathcal L_{\mathrm{GRO}}(\theta;\bar\theta)
=-\mathbb E_{(I,\mathcal A_I)\sim\mathcal D}
 \mathbb E_{y^{(1:K)}\sim\pi_{\bar\theta}}
 \left[\frac1K\sum_{j=1}^{K}
 \operatorname{sg}(A_j)\log\pi_\theta(y^{(j)}\mid I,p)\right],
\label{eq:gro}
\end{equation}
where $\operatorname{sg}$ stops gradients through the reward difference
and the update is evaluated at $\theta=\bar\theta$.
Unlike a target-selection rule, this objective retains the sign and
magnitude of each sampled answer's relative quality. Its loss terms
encourage higher-quality sequences and discourage inferior ones.
The reward is assigned to the joint framing decision, while the full
sequence probability propagates that feedback through the coordinate
answer. The greedy output serves as the reference and receives no
separate imitation loss for its coordinate sequence in this stage.

QOS learns
high-quality answers already reachable by the policy; GRO then uses
fresh alternatives to refine their relative likelihoods. The transition
transfers parameters, not a fixed candidate pool. A shared quality
function connects target discovery with refinement of the first answer.

\section{Experiments}
\label{sec:experiments}

\subsection{Experimental Settings}
\label{sec:experimental_settings}

\paragraph{Datasets.}
We evaluate on GAICv2~\citep{zeng2020grid}, FLMS~\citep{fang2014flms},
and FCDB~\citep{chen2017fcdb}, with 500, 500, and 345 test images,
respectively. These benchmarks offer complementary annotation formats:
GAICv2 rates multiple candidate windows, FLMS provides several reference
crops, and FCDB specifies one reference crop per image. Spatial training
uses RefCOCO and RefCOCO+~\citep{yu2016context}; crop training uses GAIC,
CPC~\citep{wei2018view}, and official FCDB training images. FLMS is excluded
from crop training, allowing us to examine whether the learned framing
preferences transfer beyond the supervised data sources.

\paragraph{Evaluation protocol.}
We consider automatic free-form cropping: each method returns one crop
without receiving a reference aspect ratio. On GAICv2, we report
$\mathrm{Acc}_{1/5}$ at IoU thresholds $0.85$ and $0.90$. A prediction is
accepted if its IoU with any top-five rated reference, including cutoff
ties, reaches the threshold. FLMS measures the maximum IoU between that
single prediction and its human references; FCDB uses the unique reference.
We also report boundary displacement error (BDE), the mean absolute
deviation of the four normalized coordinates. FLMS BDE uses the same
reference selected by maximum IoU, so overlap and boundary error describe
agreement with one matched frame. Scores are averaged over the full test
set. Acceptance evaluates agreement with highly rated candidates, while
IoU and BDE quantify retained-area overlap and boundary placement.

\vspace{-1.2mm}
\paragraph{Implementation details.}
\method{} uses Qwen3.5-0.8B with a frozen visual encoder. Stage I adapts
the visual merger and language LoRA; QOS and GRO update only the
language adapters. Both online stages sample four alternatives to the
current greedy answer and share an annotation-defined reward.
We use AdamW~\citep{loshchilov2019adamw} with an image batch size of 32.
The final model receives one pass of QOS followed by one pass of GRO;
their learning-rate schedules use base rates of $2\times10^{-6}$ and
$10^{-6}$, respectively. The online adapters use LoRA rank $64$ and
$\alpha=128$, and preprocessing preserves image aspect ratios.
All benchmarks use the same final checkpoint, fixed crop prompt, and
greedy coordinate decoding, with no candidate search or reranking.

\vspace{-1.5mm}
\subsection{Comparison with Existing Methods}
\label{sec:comparison}

We compare with VFN~\citep{chen2017vfn}, VEN and VPN~\citep{wei2018view},
GAICD~\citep{zeng2019grid,zeng2020grid},
CACNet~\citep{hong2021composing}, Cropper~\citep{lee2025cropper},
InstructCrop~\citep{sheng2025instructcrop}, and
Venus~\citep{du2026venus}. These methods span candidate scoring,
composition-aware prediction, and retrieval-augmented or direct VLM cropping.

\begin{table}[t]
\vspace{-3mm}
\centering
\small
\caption{\textbf{Comparison with representative cropping methods.}
Acc denotes GAICv2 $\mathrm{Acc}_{1/5}$. BDE penalizes invalid predictions
with $1$. Bold/underline mark the best/second-best results for each metric across the displayed methods evaluated under the same benchmark protocol.}
\label{tab:main_comparison}
\setlength{\tabcolsep}{2.4pt}
\renewcommand{\arraystretch}{1.05}
\begin{tabular*}{\linewidth}{@{\extracolsep{\fill}}llrrrrrr@{}}
\toprule
\multirow{2}{*}{Method} & \multirow{2}{*}{Venue}
& \multicolumn{2}{c}{GAICv2} & \multicolumn{2}{c}{FLMS}
& \multicolumn{2}{c}{FCDB}\\
\cmidrule(lr){3-4}\cmidrule(lr){5-6}\cmidrule(l){7-8}
& & $\mathrm{Acc}^{.85}\!\uparrow$ & $\mathrm{Acc}^{.90}\!\uparrow$
& IoU$\uparrow$ & BDE$\downarrow$ & IoU$\uparrow$ & BDE$\downarrow$\\
\midrule
\multicolumn{8}{@{}l}{\textit{Traditional methods}}\\[1pt]
VFN & MM'17 & 0.5500 & 0.2640 & 0.7976 & 0.0500 & 0.6433 & 0.0948\\
VEN & CVPR'18 & 0.0340 & 0.0120 & 0.7075 & 0.0619 & 0.5863 & 0.1057\\
VPN & CVPR'18 & 0.0380 & 0.0080 & 0.8234 & 0.0399 & 0.6574 & 0.0878\\
GAICD & TPAMI'22 & 0.8300 & \underline{0.7640} & 0.8025 & 0.0470 & 0.6759 & 0.0819\\
CACNet & CVPR'21 & 0.5100 & 0.2360 & 0.8401 & 0.0368 & \underline{0.6967} & \underline{0.0751}\\
\midrule
\multicolumn{8}{@{}l}{\textit{VLM-based methods}}\\[1pt]
Cropper & CVPR'25 & 0.6900 & 0.6460 & 0.7845 & 0.0829 & 0.6519 & 0.1089\\
InstructCrop & MM'25 & 0.4740 & 0.1680 & 0.8145 & 0.0422 & 0.6848 & 0.0774\\
Venus & CVPR'26 & \underline{0.8380} & 0.7200 & \underline{0.8638} & \underline{0.0304} & 0.6836 & 0.0794\\
\midrule
\method{} & Ours & \textbf{0.9140} & \textbf{0.8340} & \textbf{0.8679} & \textbf{0.0297} & \textbf{0.7124} & \textbf{0.0704}\\
\bottomrule
\end{tabular*}
\vspace{-3mm}
\end{table}

\input{figs/qualitative_comparison}

\vspace{-1.2mm}
\paragraph{Quantitative results.}
The same \method{} model leads the displayed methods in acceptance,
overlap, and boundary accuracy (Table~\ref{tab:main_comparison}).
GAICv2 acceptance improves by 7.6 and 7.0 percentage points over the
strongest listed competitors. Relative to Venus on FLMS,
BDE decreases by 2.4\%, alongside higher IoU.
The BDE ranking also holds on the common valid-image sets, indicating
closer boundary placement beyond the effect of invalid predictions.

\vspace{-1.2mm}
\paragraph{Qualitative results.}
In Figure~\ref{fig:qualitative}(a), \method{} preserves the curved
roofline while reducing foreground rocks. In (b), it balances the
mountain range and its reflection, removing excess sky and open water.
In (c), it retains the pedestrian's full body and feet, which several
competing crops truncate. These examples illustrate content-dependent
framing: reducing peripheral regions for architecture and landscapes
while retaining the spatial extent needed for an intact human subject.

\vspace{-1.5mm}
\subsection{Ablation Studies}
\label{sec:ablations}

We examine Stage-I initialization and the gains from QOS and GRO,
then compare matched objectives to assess quality-based target selection
and relative-quality refinement.

\begin{table}[t]
\vspace{-2mm}
\centering
\small
\caption{\textbf{Contributions of the three training stages.}
The upper block compares online paths from full Stage I; the lower
block varies Stage-I components, followed by matched complete-frame
supervision, QOS, and GRO. Each online stage uses one pass.
Bold is determined before rounding.}
\label{tab:training_stages}
\setlength{\tabcolsep}{2.4pt}
\renewcommand{\arraystretch}{1.1}
\begin{tabular*}{\linewidth}{@{\extracolsep{\fill}}lrrrrrr@{}}
\toprule
\multirow{2}{*}{Training configuration}
& \multicolumn{2}{c}{GAICv2} & \multicolumn{2}{c}{FLMS}
& \multicolumn{2}{c}{FCDB}\\
\cmidrule(lr){2-3}\cmidrule(lr){4-5}\cmidrule(l){6-7}
& $\mathrm{Acc}^{.85}\!\uparrow$ & $\mathrm{Acc}^{.90}\!\uparrow$
& IoU$\uparrow$ & BDE$\downarrow$ & IoU$\uparrow$ & BDE$\downarrow$\\
\midrule
Stage I & 0.9040 & 0.7840 & 0.8512 & 0.0338 & 0.7013 & 0.0745\\
Stage I $\to$ GRO & 0.9060 & \textbf{0.8340} & 0.8635 & 0.0302 & 0.7110 & 0.0704\\
Stage I $\to$ QOS & \textbf{0.9140} & 0.8320 & 0.8673 & 0.0299 & 0.7069 & 0.0722\\
Stage I $\to$ QOS $\to$ GRO & \textbf{0.9140} & \textbf{0.8340} & \textbf{0.8679} & \textbf{0.0297} & \textbf{0.7124} & \textbf{0.0704}\\
\midrule
Localization omitted & 0.8800 & 0.8120 & 0.8553 & 0.0320 & 0.7096 & 0.0717\\
Anchor supervision only & 0.9060 & 0.8300 & 0.8627 & 0.0307 & 0.7055 & 0.0729\\
\bottomrule
\end{tabular*}
\vspace{-3mm}
\end{table}

\vspace{-1.2mm}
\paragraph{Spatial and composition training.}
One variant omits localization; the other retains it but uses only anchor
supervision for composition learning, jointly removing the listwise and
high-quality-set terms. Both receive matched complete-frame supervision,
QOS, and GRO. Localization omission lowers GAICv2 acceptance by 3.4/2.2
percentage points; anchor-only supervision lowers it by 0.8/0.4 points
and reduces FCDB IoU by 0.0069. Both variants underperform the full model on all six final metrics
(Table~\ref{tab:training_stages}). These final gaps remain despite later
training, supporting spatial grounding and multi-candidate preferences.
Stagewise results are provided in the supplementary material.

\vspace{-1.2mm}
\paragraph{Contributions of QOS and GRO.}
Starting from the full Stage-I model, either online stage improves direct
prediction without changing the architecture or decoding
(Table~\ref{tab:training_stages}). QOS lowers FLMS BDE from 0.0338 to
0.0299, improving boundary placement alongside overlap. QOS alone
performs better on FLMS, whereas GRO alone gives stronger FCDB agreement.
Applying QOS followed by GRO achieves the best IoU on both datasets.
These gains motivate comparing winner imitation with relative-quality
refinement under matched training conditions.

\begin{table}[t]
\vspace{-3mm}
\centering
\small
\caption{\textbf{Target selection and policy refinement.}
(a) Models trained on different supervision targets undergo the same
GRO refinement; random selection and QOS use equal candidate budgets.
(b) Different update rules refine a common QOS model. Image exposure
and learning-rate schedules are matched within each block.
Bold marks the best values before rounding.}
\label{tab:matched_objectives}
\setlength{\tabcolsep}{2.4pt}
\renewcommand{\arraystretch}{1.1}
\begin{tabular*}{\linewidth}{@{\extracolsep{\fill}}lrrrrrr@{}}
\toprule
\multirow{2}{*}{Training objective}
& \multicolumn{2}{c}{GAICv2} & \multicolumn{2}{c}{FLMS}
& \multicolumn{2}{c}{FCDB}\\
\cmidrule(lr){2-3}\cmidrule(lr){4-5}\cmidrule(l){6-7}
& $\mathrm{Acc}^{.85}\!\uparrow$ & $\mathrm{Acc}^{.90}\!\uparrow$
& IoU$\uparrow$ & BDE$\downarrow$ & IoU$\uparrow$ & BDE$\downarrow$\\
\midrule
\multicolumn{7}{l}{\textit{(a) Target selection before policy refinement}}\\
Reference targets $\to$ GRO & 0.8920 & 0.8300 & 0.8580 & 0.0324 & 0.7109 & 0.0719\\
Random targets $\to$ GRO & 0.8560 & 0.8200 & 0.8554 & 0.0334 & 0.6997 & 0.0762\\
QOS $\to$ GRO & \textbf{0.9140} & \textbf{0.8340} & \textbf{0.8679} & \textbf{0.0297} & \textbf{0.7124} & \textbf{0.0704}\\
\midrule
\multicolumn{7}{l}{\textit{(b) Policy refinement after QOS}}\\
Continue QOS & 0.9100 & 0.8260 & 0.8657 & 0.0301 & 0.7067 & 0.0721\\
Positive-only GRO & 0.9100 & 0.8120 & 0.8661 & \textbf{0.0297} & 0.7085 & 0.0709\\
Full GRO & \textbf{0.9140} & \textbf{0.8340} & \textbf{0.8679} & 0.0297 & \textbf{0.7124} & \textbf{0.0704}\\
\bottomrule
\end{tabular*}
\vspace{-2mm}
\end{table}

\vspace{-1.2mm}
\paragraph{Quality-based target selection.}
To separate exploration from target selection, we replace the QOS winner
with a uniform random candidate while preserving the sampling budget and
online refreshing, then apply the same GRO schedule. QOS yields lower
BDE at the final endpoint: 0.0297 versus 0.0334 on FLMS, and 0.0704 versus
0.0762 on FCDB (Table~\ref{tab:matched_objectives}a). Its benefit extends
beyond crossing acceptance thresholds to more accurate boundary placement.
QOS also improves all six final metrics over fixed-reference supervision
under the same GRO refinement.

Fixed references express human preferences, while random targets follow
the current policy without favoring better framing. QOS combines both
through quality-based selection. Random supervision produces only 24
distinct FLMS boxes across 500 images, all sharing the same left boundary
and the same right boundary. GRO increases this count to 81 but does not recover QOS's
quality. This suggests that preferring better generated answers shapes
the candidates available for later policy refinement under the same
annotation-defined quality criterion.

\begin{figure}[t]
  \centering
  \includegraphics[width=\linewidth,trim=0 9pt 0 0,clip]{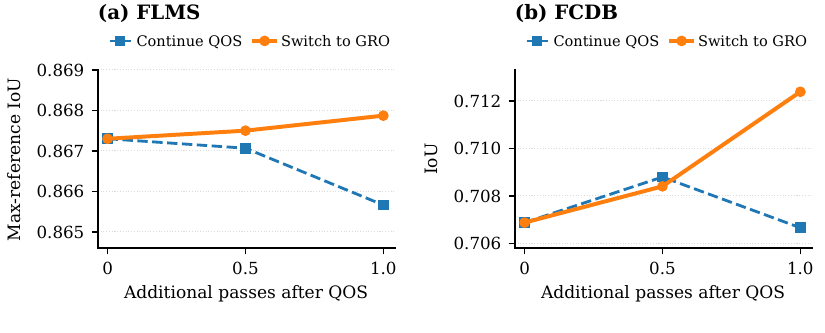}
  \vspace{-4mm}
  \caption{\textbf{Refining the QOS policy.}
  Both curves start from the same QOS model with matched image exposure
  and learning-rate schedules. Markers show initial, half-pass, and
  one-pass evaluations; lines connect measured greedy-output scores.
  The two panels use different vertical ranges.}
  \label{fig:online_refinement}
  \vspace{-2mm}
\end{figure}

\vspace{-1.2mm}
\paragraph{Relative-quality refinement.}
From the same QOS model, GRO increases FLMS IoU by about 0.0006 and
FCDB IoU by 0.0055, while reducing BDE on both datasets. Continued QOS
instead lowers final FLMS quality and loses its intermediate FCDB gain
(Figure~\ref{fig:online_refinement}). These results support switching from
winner imitation to signed quality feedback after QOS has learned to
favor high-quality complete answers among its own generated cropping alternatives.

Clipping advantages to $\max(A_j,0)$ removes feedback from worse answers.
Full GRO gives higher acceptance and IoU, and lower FCDB BDE, while
positive-only GRO has marginally lower FLMS BDE before rounding
(Table~\ref{tab:matched_objectives}b). Because clipping also changes update
magnitude, these results favor the complete rule without establishing
negative feedback as independently necessary.

\vspace{-2mm}
\section{Conclusion}
\label{sec:conclusion}
\vspace{-1.5mm}

We propose \method{}, a compact vision--language model for direct
aesthetic cropping. Spatial and composition training builds its initial
capabilities, QOS identifies high-quality answers to imitate, and GRO
refines their quality relative to the current greedy prediction.
Experiments show strong cropping performance with a sub-billion-parameter
backbone. Controlled comparisons support quality selection beyond online
sampling alone and the subsequent use of relative-quality feedback.
This training process strengthens direct decisions without adding
search or scoring at deployment. Personalized preferences and broader
photographic intentions offer promising extensions for adapting crop decisions
to individual users and diverse capture contexts.

\clearpage
\setlength{\parskip}{6pt plus 1.8pt minus 0.8pt}
\setlength{\bibsep}{11.2pt plus 2.5pt minus 0.5pt}
\bibliography{references}
\bibliographystyle{framelite_preprint}

\end{document}

%% file: figs/qualitative_comparison.tex
\begin{figure}[t]
\centering
\begingroup
\setlength{\tabcolsep}{1.1pt}
\renewcommand{\arraystretch}{1.0}
\newlength{\qualcolwidth}
\setlength{\qualcolwidth}{\dimexpr(\linewidth-14\tabcolsep)/8\relax}
\newcommand{\qualimage}[3]{%
  \begin{minipage}[c][#2][c]{\qualcolwidth}%
    \centering\includegraphics[width=\linewidth,height=#2,keepaspectratio]{figs/qualitative/#1/#3}%
  \end{minipage}}
\newcommand{\qualhead}[1]{\makebox[\qualcolwidth][c]{\scriptsize #1}}
\newcommand{\qualscore}[1]{{\scriptsize #1}}
\begin{tabular}{@{}*{8}{c}@{}}
\qualhead{Input} & \qualhead{GAICD} & \qualhead{CACNet}
& \qualhead{Cropper} & \qualhead{InstructCrop}
& \qualhead{Venus} & \qualhead{Human ref.} & \qualhead{\textbf{FrameLite}}\\[3pt]
\qualimage{curved_architecture}{36pt}{input.jpg} &
\qualimage{curved_architecture}{36pt}{gaicd.jpg} &
\qualimage{curved_architecture}{36pt}{cacnet.jpg} &
\qualimage{curved_architecture}{36pt}{cropper.jpg} &
\qualimage{curved_architecture}{36pt}{instructcrop.jpg} &
\qualimage{curved_architecture}{36pt}{venus.jpg} &
\qualimage{curved_architecture}{36pt}{human_reference.jpg} &
\qualimage{curved_architecture}{36pt}{framelite.jpg}\\[1pt]
\qualscore{(a)} & \qualscore{0.8794} & \qualscore{0.8652} & \qualscore{0.8794} & \qualscore{0.8444} & \qualscore{0.8333} & \qualscore{--} & \qualscore{\textbf{0.9253}}\\[5pt]

\qualimage{mountain_reflection}{54pt}{input.jpg} &
\qualimage{mountain_reflection}{54pt}{gaicd.jpg} &
\qualimage{mountain_reflection}{54pt}{cacnet.jpg} &
\qualimage{mountain_reflection}{54pt}{cropper.jpg} &
\qualimage{mountain_reflection}{54pt}{instructcrop.jpg} &
\qualimage{mountain_reflection}{54pt}{venus.jpg} &
\qualimage{mountain_reflection}{54pt}{human_reference.jpg} &
\qualimage{mountain_reflection}{54pt}{framelite.jpg}\\[1pt]
\qualscore{(b)} & \qualscore{0.8249} & \qualscore{0.8463} & \qualscore{0.7542} & \qualscore{0.8300} & \qualscore{0.8192} & \qualscore{--} & \qualscore{\textbf{0.8756}}\\[5pt]

\qualimage{arcade_pedestrian}{42pt}{input.jpg} &
\qualimage{arcade_pedestrian}{42pt}{gaicd.jpg} &
\qualimage{arcade_pedestrian}{42pt}{cacnet.jpg} &
\qualimage{arcade_pedestrian}{42pt}{cropper.jpg} &
\qualimage{arcade_pedestrian}{42pt}{instructcrop.jpg} &
\qualimage{arcade_pedestrian}{42pt}{venus.jpg} &
\qualimage{arcade_pedestrian}{42pt}{human_reference.jpg} &
\qualimage{arcade_pedestrian}{42pt}{framelite.jpg}\\[1pt]
\qualscore{(c)} & \qualscore{0.8308} & \qualscore{0.7960} & \qualscore{0.8340} & \qualscore{0.7566} & \qualscore{0.7966} & \qualscore{--} & \qualscore{\textbf{0.8841}}\\

\end{tabular}
\endgroup
\vspace{-3mm}
\caption{\textbf{Qualitative comparison.}
Scores below predictions are IoU. For multi-reference evaluation,
the displayed human reference may differ from each prediction's matched reference.}
\label{fig:qualitative}
\vspace{-3mm}
\end{figure}